\documentclass[conference]{IEEEtran}
\IEEEoverridecommandlockouts
\usepackage{cite}
\usepackage{amsmath,amssymb,amsfonts}
\usepackage{algorithmic}
\usepackage{graphicx}
\usepackage{textcomp}
\usepackage{xcolor}
\usepackage{soul}
\usepackage{booktabs,arydshln,graphicx,multirow}
\usepackage{caption} 
\usepackage{hyperref}

\definecolor{imageMetric}{HTML}{4F587D}
\definecolor{tempMetric}{HTML}{657F49}
\definecolor{compMetric}{HTML}{C42536}

\def\BibTeX{{\rm B\kern-.05em{\sc i\kern-.025em b}\kern-.08em
    T\kern-.1667em\lower.7ex\hbox{E}\kern-.125emX}}

\newcommand{\method}{ST-SWD}
\begin{document}

\title{Perceptual Distortion for Overfitted Video Coding:\\Sliced Matching in Space, Error Alignment in Time
    \thanks{This work is funded by the University of Bristol Research Internship programme, and the UKRI MyWorld
        Strength in Places Programme (SIPF00006/1). The authors also
        acknowledge the use of resources provided by the Isambard-AI National AI
        Research Resource (AIRR). Isambard-AI is operated by the University of
        Bristol and is funded by the UK Government's Department for Science, Innovation
        and Technology (DSIT) via UK Research and Innovation and the Science and
        Technology Facilities Council [ST/AIRR/I-A-I/1023].}
}

\author{\IEEEauthorblockN{Tianyi Yang, Ge Gao, Fan Zhang, David Bull}
    \IEEEauthorblockA{\textit{Visual Information Laboratory, University of Bristol}, Bristol, UK \\
        \{ty.yang, ge1.gao, fan.zhang, dave.bull\}@bristol.ac.uk}
}

\maketitle

\begin{abstract}
    Recent works in neural video coding have increasingly been trading off perceptual quality against reconstruction fidelity at a fixed bitrate. Wasserstein Distortion (WD) offers a principled approach to balancing these objectives by relaxing spatial correspondences in feature matching, enabling lightweight overfitted image codecs to compete with computationally intensive generative decoders. Extending vanilla WD to video, however, presents two challenges: its independent Gaussian approximation limits the expressiveness of feature distribution matching, while unconstrained texture resampling across frames introduces temporal inconsistencies. We address these challenges by distinguishing between the correspondences that can be relaxed and those that must be preserved. Spatially, we introduce a local sliced Wasserstein distortion (SWD) that matches empirical feature distributions through one-dimensional optimal transport, capturing cross-channel dependencies without enforcing exact spatial alignment. Temporally, we employ a novel flow residual that constrains coding errors along motion trajectories, preserving temporal consistency without imposing additional reconstruction fidelity constraints. Evaluated on UVG and HEVC Class~B with the overfitted codec COOL-CHIC, our framework achieves BD-rate reductions of 9.2\% in DISTS and 3.0\% in LPIPS relative to frame-wise WD, averaged over test sequences, while lowering tOF by 10.2\% at matched rate.
\end{abstract}


\section{Introduction}

Video compression is essential for the efficient transmission and storage of digital video~\cite{bull2021intelligent}. Learned video codecs~\cite{li2023dcvcdc,jia2025dcvcrt} now rival H.266/VVC~\cite{bross2021vvc} in rate--distortion performance, while overfitted codecs~\cite{dupont2021coin,kim2024c3,ladune2023coolchic,kwan2024nvrc} train a lightweight per-video decoder that cuts decoding complexity by orders of magnitude. Both, however, are mostly optimized for pixel-wise distortions such as MSE, which correlate poorly with human perception and yield blurry reconstructions at low rates.

The rate--distortion--perception theory~\cite{blau2019rdp} shows that, at a fixed rate, perceptual quality can only be improved at the expense of reconstruction fidelity. Generative video codecs exploit this trade-off with GAN~\cite{yang2022plvc} or diffusion~\cite{mao2026gnvcvd} decoders that synthesize realistic textures, but at a decoding cost incompatible with lightweight overfitted codecs.

Inspired by models of human visual perception, Wasserstein Distortion (WD)~\cite{qiu2023wd} unifies fidelity and realism by measuring the optimal-transport distance between local distributions of deep features. By allowing spatial rearrangements of perceptually similar textures, WD relaxes the requirement for exact correspondence while preserving local feature statistics, providing a perceptually motivated approach to navigating this trade-off. Ballé et al.~\cite{balle2025good} demonstrated that C3~\cite{kim2024c3}, an overfitted image codec, optimized with WD rivals generative codecs in perceptual quality at a fraction of decoding cost.

Despite its effectiveness, existing implementations of WD rely on a simplified approximation of the Wasserstein distance. Specifically, local feature distributions are modeled as independent Gaussians, reducing distribution matching to comparisons of per-channel means and standard deviations. While computationally efficient, this approximation overlooks higher-order statistics and cross-channel dependencies, potentially allowing perceptually distinct textures to incur similar distortion penalties.

Extending WD to video introduces a further challenge. By matching feature distributions independently for each frame, WD allows spatial texture rearrangements without enforcing consistency across time. Consequently, the same scene content may be reconstructed with different texture realizations across successive frames, leading to noticeable temporal flickering and instability. This reveals a fundamental distinction between spatial and temporal correspondences: while exact spatial alignment can be relaxed for perceptually plausible texture reconstruction, temporal coherence requires consistency along motion trajectories.

\begin{figure*}[t]
    \centering
    \includegraphics[width=\textwidth]{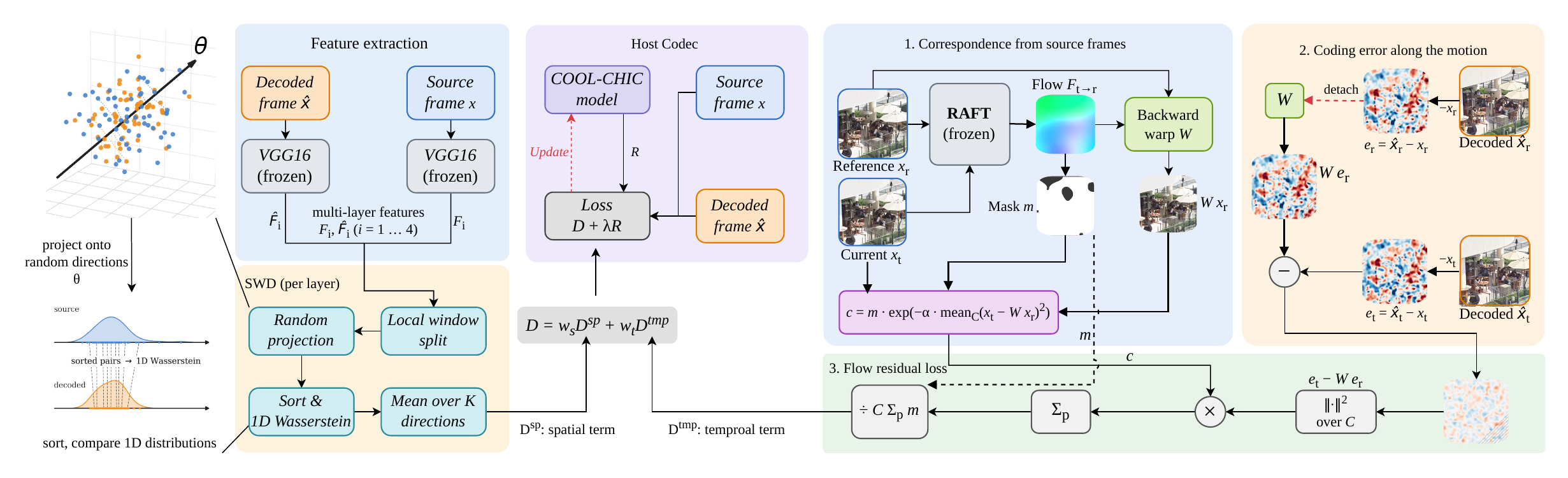}
    \caption{\textbf{Overview of \method{}.} The host codec (COOL-CHIC) is optimized for $D+\lambda R$ (center), where $D$ comprises the local sliced Wasserstein distortion $D^{\mathrm{sp}}$ (left) and the flow residual $D^{\mathrm{tmp}}$ (right).}
    \label{fig:overview}
\end{figure*}

To address these issues, we propose \textbf{Spatio-Temporal Sliced WD (\method{})}, a novel perceptual objective for overfitted video coding that tightens WD within frames and anchors it across frames. First, we replace the moment-based approximation with a \emph{local sliced Wasserstein distortion} (SWD). Rather than comparing per-channel moments, SWD projects local feature vectors onto random directions and compares their sorted projections, performing exact one-dimensional optimal transport along each direction to capture richer distributional statistics, including cross-channel dependencies. Second, we introduce a \emph{flow residual} that penalizes changes in coding errors along motion trajectories estimated from the source frames. By constraining motion-aligned coding errors rather than their magnitude, this term encourages temporally consistent texture reconstruction and suppresses flickering. We integrate \method{} into the overfitted video codec COOL-CHIC~5.0~\cite{ladune2026coolchic5} as a drop-in replacement for WD, leaving the decoder and bitstream unchanged. Across UVG and HEVC Class~B, \method{} achieves average BD-rate savings of \textbf{3.0\%} and \textbf{9.2\%} when evaluated using the perceptual metrics LPIPS and DISTS, respectively, while reducing the temporal metric tOF by \textbf{10.2\%} relative to frame-wise WD.

\section{Method}

\subsection{Preliminaries}
\label{sec:preliminaries}
Let $x$ and $\hat{x}$ denote a source video and its reconstruction. Overfitted codecs optimize instance-specific parameters $\theta$ by minimizing the rate--distortion trade-off $D(x,\hat{x}) + \lambda R(\theta)$, where $R$ denotes the bitrate; depending on the codec, the coding unit may be a frame~\cite{ladune2023coolchic,kim2024c3} or a group of frames or patches~\cite{kwan2023hinerv,kwan2024nvrc}. We adopt COOL-CHIC~5.0~\cite{ladune2026coolchic5,coolchicsoftware} as the host codec, as it is a state-of-the-art open-source overfitted codec with inter-frame coding, and its released encoder supports WD for still images~\cite{philippe2025clic}, which we extend to video as our WD baseline (Sec.~\ref{sec:setup}). COOL-CHIC encodes each frame $x_t$ in coding order over frame-specific parameters $\theta_t$. For inter frames, reconstruction combines motion-compensated prediction from previously decoded reference frames, which remain fixed during the optimization of the current frame, with a learned residual~\cite{coolchicvideo,leguay2025improved,ladune2025subpixel}. Its distortion is $D = w_{\mathrm{m}}\,\mathrm{MSE} + w_{\mathrm{s}}\,D^{\mathrm{sp}}$, where the two terms correspond to a fidelity anchor and a spatial perceptual distortion, respectively, and the deployed image configuration sets $D^{\mathrm{sp}}$ to a rescaled WD.

The deployed WD~\cite{balle2025good,philippe2025clic} extracts features from four stages of VGG-16~\cite{simonyan2015vgg} and independently computes local means $\mu_{s,i}$ and standard deviations $\sigma_{s,i}$ for each feature channel at pooling level $i$. The distortion is then defined as
\begin{equation}
    \sum_{s}\sum_{i} w_i\big[(\mu_{s,i}-\hat{\mu}_{s,i})^2 + (\sigma_{s,i}-\hat{\sigma}_{s,i})^2\big],
    \label{eq:wd}
\end{equation}
summed over stages $s$ and pooling levels $i$, where the weights $w_i$ depend on the pooling scale. This scale is fixed by default, leaving only one active pooling level per stage.

\subsection{Overview}
Fig.~\ref{fig:overview} illustrates \method{}, which modifies only the distortion objective of the host codec. The spatial term $D^{\mathrm{sp}}$ replaces the deployed WD with a local sliced Wasserstein distortion: VGG-16 features of the source and reconstructed frames are projected onto random directions within local windows and compared through one-dimensional optimal transport, capturing richer local statistics while allowing spatial rearrangements of perceptually similar textures. The temporal term $D^{\mathrm{tmp}}$ is a flow residual: optical flow estimated from the source frames by RAFT~\cite{teed2020raft} aligns the coding errors of the current and reference frames, and their difference is penalized, encouraging temporally coherent reconstructions without directly penalizing the error magnitude. The overall distortion is
\begin{equation}
    D = w_{\mathrm{m}}\,\mathrm{MSE} + w_{\mathrm{s}}\,D^{\mathrm{sp}} + w_{\mathrm{t}}\,D^{\mathrm{tmp}}.
    \label{eq:total}
\end{equation}

\subsection{Local sliced Wasserstein distortion}
\label{sec:swd}
The sliced Wasserstein distance approximates high-dimensional optimal transport by averaging one-dimensional Wasserstein distances over random projections~\cite{rabin2011swd,bonneel2015swd}. Heitz et al.~\cite{heitz2021swd} demonstrated its effectiveness for deep-feature texture synthesis, where it captures distributional information beyond the first two moments. Let $P, \hat{P} \in \mathbb{R}^{N\times K}$ denote the feature matrices of the source and reconstruction, respectively, where each row corresponds to a $K$-dimensional feature vector (i.e., channel dimension) and $N$ is the number of spatial positions. Given $M$ unit projection directions $\mathbf{u}_k$ sampled uniformly from $\mathbb{S}^{K-1}$ and shared between the source and reconstruction, we define
\begin{equation}
    \mathrm{SWD}(P,\hat{P}) = \frac{1}{M}\sum_{k=1}^{M}\big\lVert \mathrm{sort}(P\mathbf{u}_k) - \mathrm{sort}(\hat{P}\mathbf{u}_k)\big\rVert_2^2.
    \label{eq:swd}
\end{equation}
Sorting yields the optimal one-dimensional transport matching~\cite{bonneel2015swd}. Compared to the moment-based approximation of WD, SWD captures richer distributional statistics, including cross-channel dependencies through projections that jointly combine feature channels.

Existing applications of SWD typically match feature distributions over the entire image~\cite{heitz2021swd,elnekave2022gpdm}, which is suitable for texture synthesis but provides insufficient spatial localization for compression. We therefore restrict distribution matching to local windows: at the same four VGG-16 stages as~\autoref{eq:wd}, $P$ collects feature vectors within non-overlapping $16\times16$ windows. The window grid is randomly shifted at each optimization step to prevent artifacts associated with fixed boundaries. At each stage, \autoref{eq:swd} is summed over windows and normalized by the number of feature positions, while $D^{\mathrm{sp}}$ sums equally across the four stages. Directions are resampled at each step ($M=32$), whereas evaluation and rate--distortion optimized quantization use $128$ fixed, seeded directions.

\subsection{Flow residual}
\label{sec:flowres}
Unlike spatial texture matching, temporal consistency requires correspondences along motion trajectories. We therefore constrain coding errors at corresponding positions. Let $f_{t\to r}$ and $f_{r\to t}$ denote the bidirectional optical flows estimated by RAFT-large~\cite{teed2020raft} on the \emph{source} frames. The backward warp $W$ aligns reference $r$ to current frame $t$ through $(Wz_r)(p)=z_r(p+f_{t\to r}(p))$, using bilinear sampling. Let $m(p)\in\{0,1\}$ denote the forward--backward consistency and in-frame mask in current-frame coordinates. Defining signed coding errors as $e_t=\hat{x}_t-x_t$ and $e_r=\operatorname{stopgrad}(\hat{x}_r-x_r)$, we penalize their motion-aligned difference:
\begin{equation}
    D^{\mathrm{tmp}} = \frac{\sum_p c(p)\,\lVert e_t(p) - (W e_r)(p)\rVert_2^2}{C\sum_p m(p)},
    \label{eq:flowres}
\end{equation}
where $C$ is the number of color channels and $c(p)=m(p)\exp(-\alpha\,\lVert x_t(p)-(Wx_r)(p)\rVert_2^2/C)$ is a confidence weight, with $\alpha=50$. This weight suppresses unreliable correspondences where the warped source differs from the current frame. Minimizing this term encourages $e_t=We_r$, allowing coding errors to persist along motion trajectories without directly penalizing their magnitude. Unlike the warping loss of Lai et al.~\cite{lai2018blind}, which penalizes $\lVert\hat{x}_t-W\hat{x}_r\rVert_2^2$ and directly constrains reconstructed values, our formulation penalizes temporal changes in signed coding errors. The consistency check follows~\cite{ruder2016artistic}, and errors are computed in RGB on $[0,1]$. We consider only references at a display distance of one frame and discard those with fewer than half of their pixels passing the consistency check. For B-frames, we average over the retained references. The temporal term is set to zero when no valid reference is available, including for intra frames.

\begin{figure}[t]
    \centering
    \includegraphics[width=\columnwidth]{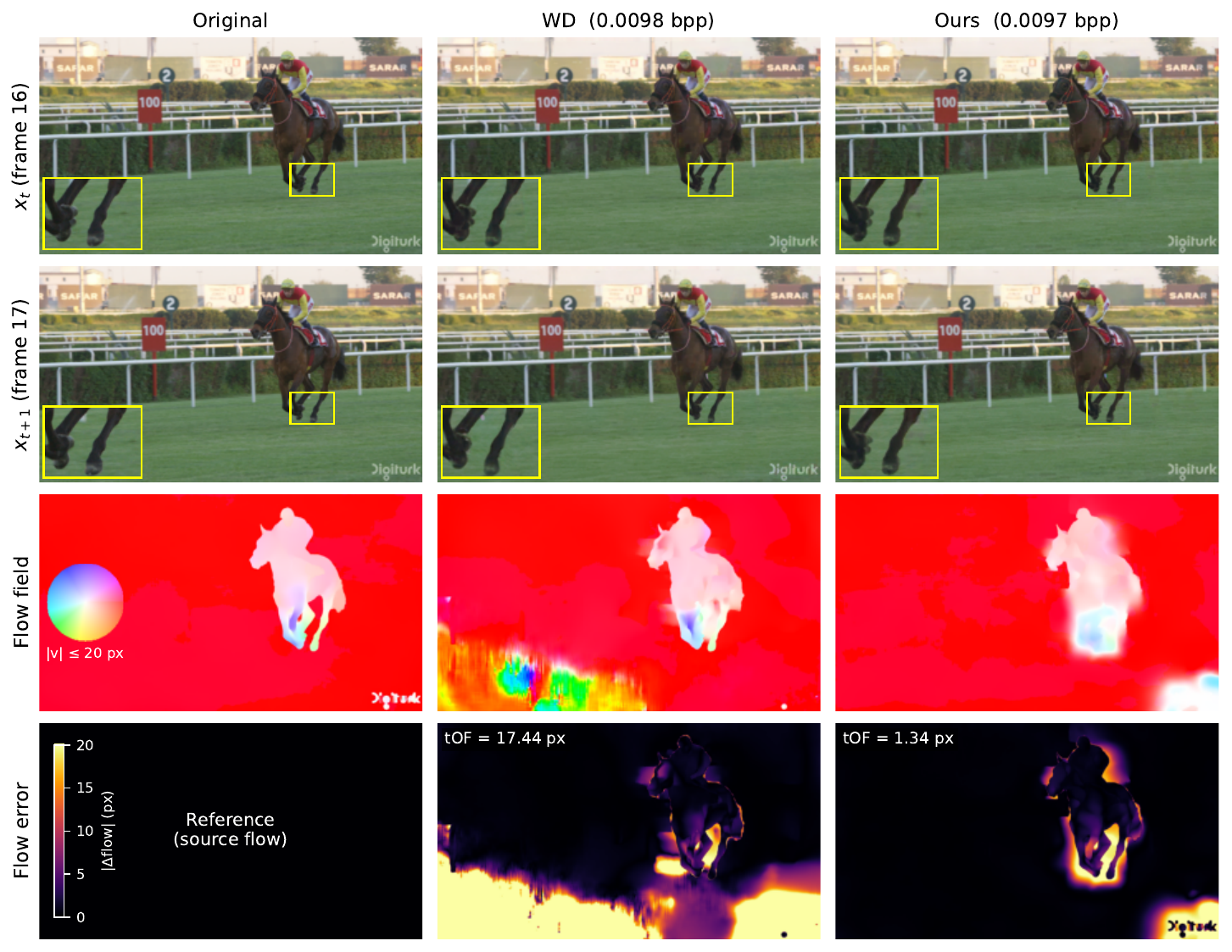}
    \caption{\textbf{Temporal consistency on Jockey (UVG)}. WD re-draws the lawn every frame, producing spurious motion; the flow residual confines the error to the horse's legs.}
    \label{fig:flow}
\end{figure}

\begin{figure*}[t]
    \centering
    \begin{minipage}[t]{0.6\textwidth}
        \vspace{0pt}
        \centering
        \captionof{table}{\textbf{Quantitative results} on UVG and HEVC-B: \method{} vs.\ each row (negative: \method{} better). \mbox{--:~$<$20\%~overlap}. Dec./Enc.: sec/frame, 1 CPU thread ($^\dagger$RTX~5090).}
        \label{tab:main}
        \setlength{\tabcolsep}{2pt}
        \renewcommand{\arraystretch}{1.0}
        \footnotesize
        \resizebox{\linewidth}{!}{%
            \begin{tabular}{l*{5}{c}cc}
                \toprule
                \textbf{Anchor}
                & \textcolor{imageMetric}{PSNR$\downarrow$}
                & \textcolor{imageMetric}{LPIPS$\downarrow$}
                & \textcolor{imageMetric}{DISTS$\downarrow$}
                & \textcolor{tempMetric}{tOF$\downarrow$}
                & \textcolor{tempMetric}{$E_{\mathrm{warp}}\!\downarrow$}
                & Dec.$\downarrow$
                & Enc.$\downarrow$ \\
                \midrule
                \multicolumn{6}{c}{\textbf{UVG}} & & \\
                \midrule
                x265 & $+36.60$\% & $-77.74$\% & -- & $-4.99$\% & $+8.59$\% & 0.003 & 1.00 \\
                VTM & $+345.82$\% & $-61.51$\% & -- & $+100.99$\% & $+49.82$\% & 0.022 & 17.6 \\
                \noalign{\vskip 2pt}\hdashline[2pt/2pt]\noalign{\vskip 2pt}
                COOL-CHIC, MSE & $+103.19$\% & $-72.66$\% & -- & $+30.14$\% & $+50.61$\% & 3.9 & 447$^\dagger$ \\
                COOL-CHIC, WD & $-0.23$\% & $-5.56$\% & $-13.20$\% & $-12.12$\% & $-8.84$\% & 4.2 & 1032$^\dagger$ \\
                \noalign{\vskip 2pt}\hdashline[2pt/2pt]\noalign{\vskip 2pt}
                \textbf{\method{} (ours)} & $\mathbf{+0.00}$\% & $\mathbf{+0.00}$\% & $\mathbf{+0.00}$\% & $\mathbf{+0.00}$\% & $\mathbf{+0.00}$\% & 4.2 & 1053$^\dagger$ \\
                \midrule
                \multicolumn{6}{c}{\textbf{HEVC-B}} & & \\
                \midrule
                x265 & $+66.72$\% & $-82.97$\% & -- & $+3.04$\% & $+9.67$\% & \multicolumn{2}{c}{\multirow{4}{*}[-2pt]{same as above}} \\
                VTM & $+401.30$\% & $-66.87$\% & -- & $+63.84$\% & $+33.47$\% & \multicolumn{2}{c}{} \\
                \noalign{\vskip 2pt}\cdashline{1-6}[2pt/2pt]\noalign{\vskip 2pt}
                COOL-CHIC, MSE & $+127.75$\% & $-76.80$\% & -- & $+18.45$\% & $+33.59$\% & \multicolumn{2}{c}{} \\
                COOL-CHIC, WD & $+3.22$\% & $+0.64$\% & $-3.62$\% & $-7.96$\% & $-5.09$\% & \multicolumn{2}{c}{} \\
                \bottomrule
            \end{tabular}}
    \end{minipage}\hfill
    \begin{minipage}[t]{0.38\textwidth}
        \vspace{0pt}
        \centering
        \includegraphics[width=\linewidth]{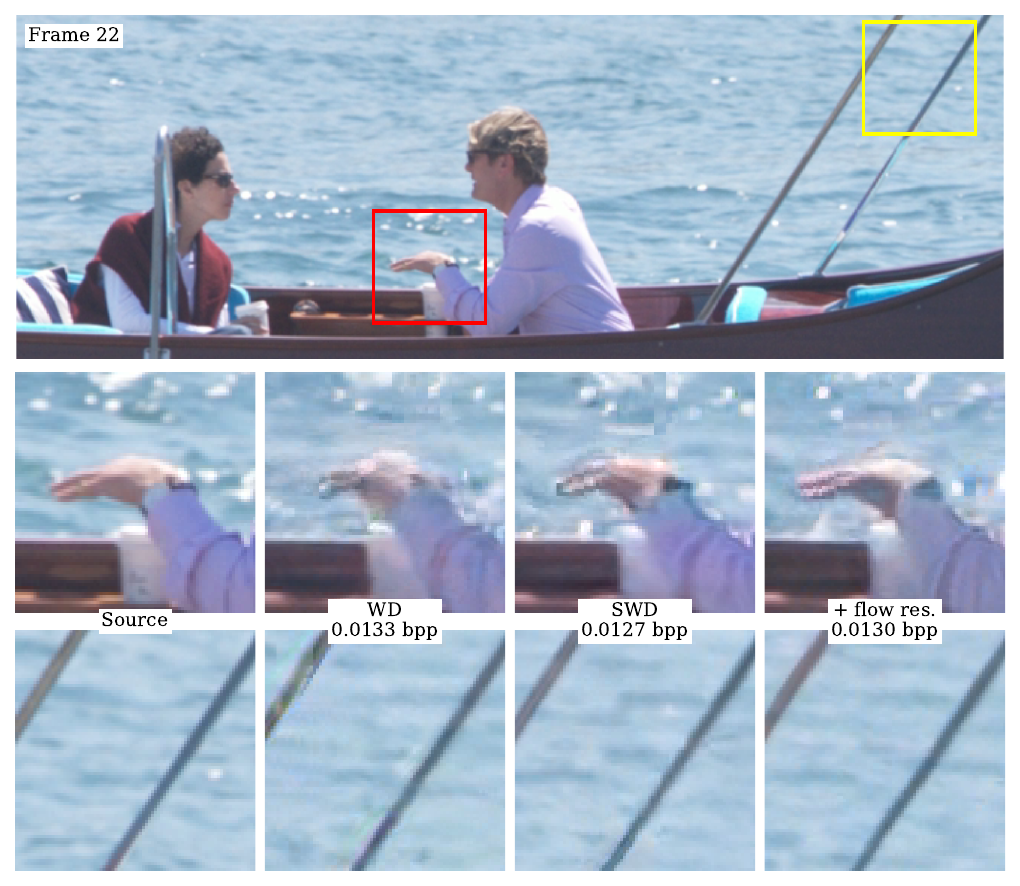}
        \captionof{figure}{\textbf{Visual comparison on YachtRide}.}
        \label{fig:crops}
    \end{minipage}
\end{figure*}

\begin{figure*}[t]
    \centering
    \begin{minipage}[t]{0.6\textwidth}
        \vspace{0pt}
        \centering
        \includegraphics[width=\linewidth]{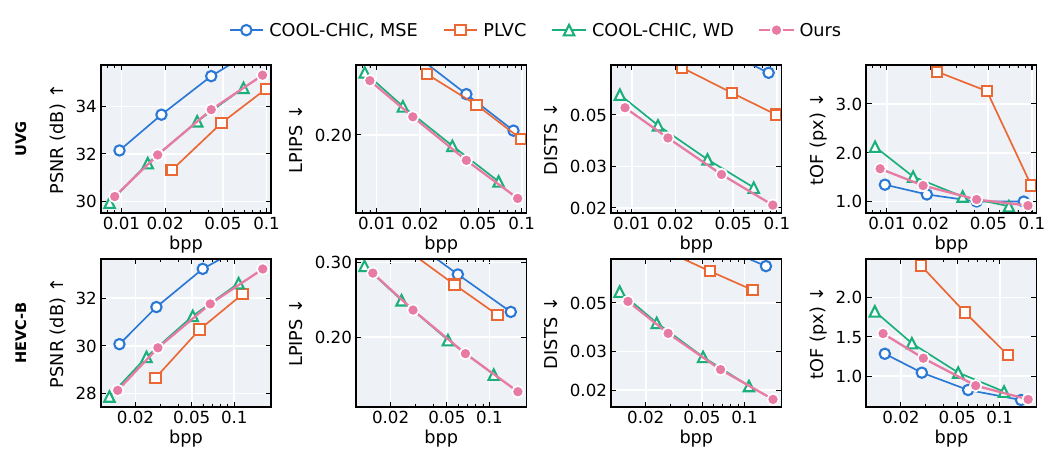}
        \captionof{figure}{\textbf{Rate--distortion curves} on UVG~\cite{mercat2020uvg} and HEVC-B~\cite{bossen2013ctc}.}
        \label{fig:rd}
    \end{minipage}\hfill
    \begin{minipage}[t]{0.38\textwidth}
        \vspace{12pt} 
        \centering
        \captionof{table}{\textbf{Ablation studies on the UVG dataset~\cite{mercat2020uvg} excluding Beauty.} Spatial metrics are BD-rates against \method{}; tOF is the mean change at matched rate.}
        \label{tab:ablation}
        \setlength{\tabcolsep}{3pt}
        \footnotesize
        \begin{tabular}{lrrr}
            \toprule
            \textbf{Metric}                         & \multicolumn{1}{c}{(\textbf{V1})} & \multicolumn{1}{c}{(\textbf{V2})} & \multicolumn{1}{c}{(\textbf{V3})} \\
            \midrule
            \textbf{\textcolor{imageMetric}{PSNR}}  & $-1.13$\%                         & $+1.10$\%                         & $+1.17$\%                         \\
            \textbf{\textcolor{imageMetric}{LPIPS}} & $+7.88$\%                         & $-2.32$\%                         & $-0.26$\%                         \\
            \textbf{\textcolor{imageMetric}{DISTS}} & $+19.03$\%                        & $-5.93$\%                         & $-0.73$\%                         \\
            \textbf{\textcolor{tempMetric}{tOF}}    & $-4.18$\%                         & $+22.02$\%                        & $+0.22$\%                         \\
            \bottomrule
        \end{tabular}
    \end{minipage}
\end{figure*}

\section{Experiments}

\subsection{Experimental Setup}
\label{sec:setup}
\noindent \textbf{Test Conditions.} We evaluate on UVG~\cite{mercat2020uvg} (7 sequences) and HEVC Class~B~\cite{bossen2013ctc} (5 sequences), both at 1080p resolution. Following the common test conditions (CTC)~\cite{bossen2013ctc}, we use a random-access (RA) configuration with an intra period of 32 frames and a hierarchical B-frame prediction structure, and following GNVC-VD~\cite{mao2026gnvcvd} we evaluate the first 96 frames of each sequence. We compare our method (\method{}) against conventional codecs x265~\cite{x265} and VTM-20.0~\cite{bross2021vvc}, COOL-CHIC~5.0~\cite{ladune2026coolchic5,coolchicsoftware} optimized with MSE and deployed WD (\autoref{eq:wd}), and PLVC~\cite{yang2022plvc}. Since the released encoder offers WD for still images only, we extend it to video by applying it frame by frame on RGB with the published MSE anchor, while the motion-estimation pre-training stage of COOL-CHIC remains MSE-based; the WD baseline and \method{} share this setup. In~\autoref{eq:total}, we set $w_{\mathrm{m}}=w_{\mathrm{t}}=0.2$ and $w_{\mathrm{s}}=0.0036$, close to the $0.8/200$ of the rescaled WD~\cite{philippe2025clic}. Following the CTC, we use QPs of $\{22, 27, 32, 37\}$ for the conventional codecs and $\lambda \in \{6\cdot10^{-5}, 2\cdot10^{-4}, 7\cdot10^{-4}, 2\cdot10^{-3}\}$ for all COOL-CHIC variants, and compare methods by BD-rate.

\noindent \textbf{Metrics.} Following GNVC-VD~\cite{mao2026gnvcvd}, we measure bitrate in bits per pixel (bpp) and evaluate per-frame quality using PSNR, LPIPS~\cite{zhang2018lpips}, and DISTS~\cite{ding2020dists}. For temporal fidelity, we report tOF~\cite{chu2020tecogan}, which measures the discrepancy in optical flow between the source and reconstruction, and the warping error $E_{\mathrm{warp}}$~\cite{lai2018blind}, computed on the reconstruction with its own RAFT flow and with occlusions masked by forward--backward consistency~\cite{ruder2016artistic}. Each row of Table~\ref{tab:main} reports \method{} relative to that method; BD-rates are omitted if quality-range overlap is below 20\% of the narrower range. Temporal metrics report per-sequence changes at matched rates, using PCHIP interpolation of the anchor in log-rate within its range. Beauty is excluded from temporal evaluation because RAFT fails; its flow residual is also disabled, reducing \method{} to SWD.

\subsection{Main results}
\label{sec:results}

\textbf{Rate-distortion performance.} As shown in Table~\ref{tab:main} and Fig.~\ref{fig:rd}, \method{} improves perceptual quality over frame-wise COOL-CHIC (WD), achieving DISTS BD-rate savings of 13.2\% on UVG and 3.6\% on HEVC-B, with PSNR BD-rate changes within 3.2\%. LPIPS also improves on UVG and remains comparable on HEVC-B. In Fig.~\ref{fig:crops}, \method{} better preserves hand details and avoids color fringing along the rigging cable. Conventional codecs and MSE-optimized COOL-CHIC achieve better PSNR and mostly lower temporal errors, but require substantially higher bitrates at matched LPIPS. PLVC is shown in Fig.~\ref{fig:rd} for reference, but its BD-rates are not reported, as only three rate points are available.

\textbf{Temporal consistency.} On UVG and HEVC-B, \method{} reduces tOF by 12.1\% and 8.0\% and $E_{\mathrm{warp}}$ by 8.8\% and 5.1\%, respectively, at matched rate. On the challenging and highly dynamic Jockey sequence (Fig.~\ref{fig:flow}), WD introduces spurious motion in the reconstructed lawn, whereas \method{} substantially reduces this artifact, lowering the frame-pair tOF from 17.4 to 1.3~px. These results are consistent with the intended role of the flow residual: encouraging coding errors to remain consistent along source motion trajectories. Sec.~\ref{sec:ablation} examines the contribution of each component and quantifies the trade-off between temporal consistency and spatial quality.

\textbf{Computational complexity.} \method{} modifies only the training objective, so its decoder and bitstream are those of the WD baseline. Decoding time is therefore unchanged (measured with COOL-CHIC's reference Python decoder, not an optimized C one, with bit-exact faster motion compensation and noise generation), while encoding time increases by approximately 2\% relative to frame-wise WD.

\subsection{Ablations}
\label{sec:ablation}

We ablate \method{} on UVG~\cite{mercat2020uvg} excluding Beauty: \textbf{(V1)} replaces SWD with the deployed WD; \textbf{(V2)} removes the flow residual; \textbf{(V3)} replaces it with the warping loss of~\cite{lai2018blind}. As Table~\ref{tab:ablation} shows, SWD accounts for most of the perceptual gain: reverting to WD increases the BD-rates, measured in DISTS and LPIPS, by 19.0\% and 7.9\%, respectively, while yielding slightly lower tOF ($-4.2$\%), as its blurrier textures vary less over time. Removing the flow residual raises tOF by 22.0\% at a cost of 5.9\% in DISTS. The warping loss obtains a similar tOF but a higher PSNR BD-rate, as it constrains reconstructed values rather than coding errors.

\section{Conclusion}
We presented \method{}, a perceptual distortion for overfitted video coding that matches richer local feature statistics than WD without requiring spatial alignment, and preserves temporal correspondence through a flow residual on coding errors. As a drop-in replacement for WD, it leaves the decoder and bitstream unchanged. Future work will remove the reliance on source optical flow and extend the temporal term beyond adjacent references.

\bibliographystyle{IEEEtran}
\bibliography{refs}

\end{document}